\pdfoutput=1

\documentclass[11pt]{article}

\usepackage{ACL2023}

\usepackage{times}
\usepackage{latexsym}

\usepackage[T1]{fontenc}

\usepackage[utf8]{inputenc}

\usepackage{microtype}

\usepackage{inconsolata}

\usepackage{subfigure}
\usepackage{graphicx}
\usepackage{wrapfig}
\usepackage{amssymb}
\usepackage{enumitem}
\usepackage{multirow}
\usepackage{booktabs}
\usepackage{amsmath}
\usepackage{float}

\title{AV-TranSpeech: Audio-Visual Robust Speech-to-Speech Translation}

\author{%
  Rongjie Huang$^{1}$\thanks{~~Equal contributions}, Huadai Liu$^{1}$\footnotemark[1], Xize Cheng$^{1}$\footnotemark[1], Yi Ren$^{2}$, Linjun Li$^{1}$, Zhenhui Ye$^{1}$, \\ \textbf{Jinzheng He$^{1}$, Lichao Zhang$^{1}$, Jinglin Liu$^{2}$, Xiang Yin$^{2}$, Zhou Zhao$^{1}$\thanks{~~Corresponding author}} \\
  Zhejiang University$^1$, ByteDance$^2$ \\ \
  \texttt{\{rongjiehuang,liuhuadai,zhaozhou\}@zju.edu.cn} \\
    \texttt{\{ren.yi,liu.jinglin,yinxiang.stephen\}@bytedance.com}
  }

\begin{document}
\maketitle
\begin{abstract}
Direct speech-to-speech translation (S2ST) aims to convert speech from one language into another, and has demonstrated significant progress to date. Despite the recent success, current S2ST models still suffer from distinct degradation in noisy environments and fail to translate visual speech (i.e., the movement of lips and teeth). In this work, we present AV-TranSpeech, the first audio-visual speech-to-speech (AV-S2ST) translation model without relying on intermediate text. AV-TranSpeech complements the audio stream with visual information to promote system robustness and opens up a host of practical applications: dictation or dubbing archival films. To mitigate the data scarcity with limited parallel AV-S2ST data, we 1) explore self-supervised pre-training with unlabeled audio-visual data to learn contextual representation, and 2) introduce cross-modal distillation with S2ST models trained on the audio-only corpus to further reduce the requirements of visual data. Experimental results on two language pairs demonstrate that AV-TranSpeech outperforms audio-only models under all settings regardless of the type of noise. With low-resource audio-visual data (10h, 30h), cross-modal distillation yields an improvement of 7.6 BLEU on average compared with baselines.\footnote{Audio samples are available at \url{https://AV-TranSpeech.github.io/}.}

\end{abstract}

\section{Introduction}
\begin{figure*}[h]
    \centering
    \includegraphics[width=1.0\textwidth,trim={1.0cm 0cm 1.0cm 0cm}]{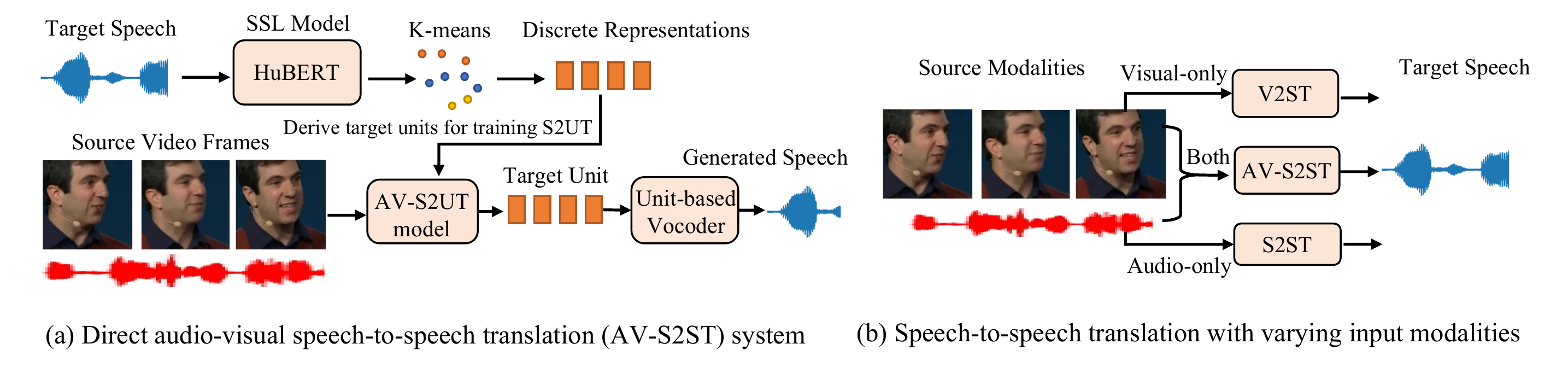}
   \caption{Beyond speech-to-speech translation (S2ST), we introduce visual-to-speech translation (V2ST) and audio-visual speech-to-speech translation (AV-S2ST), unlocking the ability for high-quality translation given a user-defined modality input. } 
    \label{fig:pipeline}
  \end{figure*}

Speech-to-speech translation (S2ST) models~\citep{tjandra2019speech,zhang2020uwspeech,jia2021translatotron} relying on speech data have achieved high performance and significantly broken down communication barriers between people not sharing a common language, which attracts broad interest in the machine learning community~\citep{huang2022prodiff,huanggenerspeech}. Among them, direct systems~\citep{lee2021direct,lee2021textless,huang2022transpeech} leverage recent progress on self-supervised discrete units learned from unlabeled speech for building textless S2ST, further supporting translation between unwritten languages~\citep{chen2022speech}.

As speech production~\citep{huang2023audiogpt,lam2021bilateral,huang2022fastdiff} is accompanied by the movement of lips and teeth, it can be visually interpreted to understand speech. In recent years, significant research~\citep{shi2022learning,prajwal2022sub} has introduced joint modeling of spoken language and vision: \citet{shi2022learning1} investigate to learn lip-based audio-visual speaker embeddings, where the speaker’s mouth area is used alongside speech as inputs. \citet{chern2022audio} focus on audio-visual speech enhancement and separation which better integrates visual information. Despite their success, it is unclear how lip can contribute to audio-based S2ST, and how to incorporate visual modality as auxiliary information in S2ST. A visual translator may open up a host of practical applications: improving speech translation in noisy environments, enabling dictation, or dubbing archival films.


Despite the benefits of audio-visual approaches, training direct speech translation models without relying on intermediate text typically requires a large amount of training data, while there are very few resources providing parallel audio-visual speech due to the heavy workload. To mitigate the data scarcity, researchers have leveraged multitask learning~\citep{lee2021direct}, data augmentation~\citep{popuri2022enhanced}, and weekly-supervised data with synthesized speech~\citep{jia2022leveraging} in audio S2ST.

In this work, we propose AV-TranSpeech, introducing the first AV-S2ST system without using text. As illustrated in Figure~\ref{fig:pipeline}, our textless AV-TranSpeech inherits speech-to-unit translation (S2UT) framework~\citep{lee2021textless,huang2022transpeech}, which consists of an audio-visual speech-to-unit translation (AV-S2UT) model followed by a unit-based vocoder that converts discrete units to speech. AV-TranSpeech complements the audio stream with the auxiliary visual information, which is invariant to speaking environments and promotes system robustness. To tackle the challenges of data shortage, we 1) build upon the recently introduced Audio-Visual HuBERT (AV-HuBERT) which learns contextual representations through self-supervised masked prediction, and show that large-scale pre-training benefits AV-S2ST training; 2) introduce cross-modal distillation and leverage S2ST models trained on the audio-only corpus, which further reduces the requirements of visual data and boosts the performance of visual systems in low-resource scenarios. 

Experimental results on two language pairs demonstrate the robustness of AV-TranSpeech in noisy scenarios, outperforming audio-only S2ST under all settings regardless of the SNR and the type of noise. With low-resource audio-visual data (10h, 30h), cross-modal distillation yields an improvement of 7.6 BLEU on average compared with baselines. The main contributions of this work include:


\begin{itemize} [leftmargin=*]
    \item We propose the first textless audio-visual speech-to-speech (AV-S2ST) translation model AV-TranSpeech and collect a benchmark dataset LRS3-T which we plan to release.
    \item We leverage the recent success of audio-visual self-supervised learning for contextual representations and show that large-scale pre-training alleviates the data scarcity issue in AV-S2ST systems.
    \item We introduce cross-modal knowledge distillation, which further reduces the requirements of visual data and boosts AV-S2ST performances in low-resource scenarios.
    \item Experimental results on two language pairs demonstrate the robustness of AV-TranSpeech in the noisy environment under all settings regardless of the type of noise, and cross-modal distillation yields a significant improvement compared with baselines.

\end{itemize}


\section{Related Work}

\subsection{Speech-to-Speech Translation}

Direct speech-to-speech translation has made massive progress to date. Translatotron~\citep{jia2019direct} is the first direct S2ST model and shows reasonable translation accuracy and speech naturalness. Translatotron 2~\citep{jia2021translatotron} utilizes the auxiliary target phoneme decoder to promote translation quality but still needs phoneme data during training. UWSpeech~\citep{zhang2020uwspeech} builds the VQ-VAE model and discards transcript in the target language, while paired speech and phoneme corpora of written language are required. Most recently, a textless S2ST system~\citep{lee2021direct} takes advantage of self-supervised learning (SSL) and leverages speech-to-unit translation (S2UT) model followed by a unit-based vocoder that converts discrete units to speech, demonstrating the results without using text data. ~\citet{popuri2022enhanced} show that self-supervised encoder and decoder pre-training with weakly-supervised data improves model performance. ~\citet{huang2022transpeech} apply speech normalization on rhythm, pitch, and energy to create deterministic training targets.

Despite their recent success, current S2ST models still suffer from distinct degradation in noisy scenarios and fail to translate visual speech. In this work, we complement the audio stream with the visual information, opening up a host of practical applications (improving speech translation in noisy environments, enabling silent dictation, or dubbing silent archival films), which has been relatively overlooked.

\subsection{Audio-Visual Speech Self-Supervised Learning}

It has been an increasing interest in self-supervised learning in the machine learning and speech-processing community. Wav2vec 2.0~\citep{baevski2020wav2vec} trains a convolutional neural network to distinguish true future samples from random distractor samples using a contrastive predictive coding (CPC) loss function. HuBERT~\citep{hsu2021hubert} is trained with a masked prediction with masked continuous audio signals. For audio-visual representation learning, they rely on spatio-temporal CNNs consisting of multiple 3D convolutional layers or a single 3D convolutional layer followed by 2D ones. ~\citet{stafylakis2017combining} propose a residual network with 3D convolutions to extract more powerful representations. 

Very recently, ~\citet{shi2022learning} propose AV-HuBERT with a self-supervised representation learning framework by masking multi-stream video input and predicting automatically discovered multimodal hidden units. It has been demonstrated to learn discriminative audio-visual speech representation, and thus we leverage the contextual representations to enhance the AV-S2ST performance.

\subsection{Transfer Learning}
Transferring knowledge across domains is a promising machine learning methodology for solving the data shortage problem. ~\citet{zhang2021transfer} perform transfer learning from a text-to-speech system to voice conversion with non-parallel training data. ~\citet{afouras2020asr} apply cross-modal distillation from ASR for learning audio-visual speech recognition, where they train strong models for visual speech recognition without requiring human annotated ground-truth data. ~\citet{cai2020speaker} enhance the knowledge transfer from the speaker verification to the speech synthesis by engaging the speaker verification network. ~\citet{popuri2022enhanced} perform transfer learning from a natural language expert mBART for faster coverage of training speech translation models. 

Our approach leverages networks trained on one modality to transfer knowledge to another. In this way, the dependence on a large number of parallel audio-visual data can be reduced for constructing AV-S2ST systems.

\section{AV-TranSpeech}

\begin{figure*}[h]
    \centering
    \includegraphics[width=1.0\textwidth,trim={0.3cm 0cm 1.6cm 0cm}]{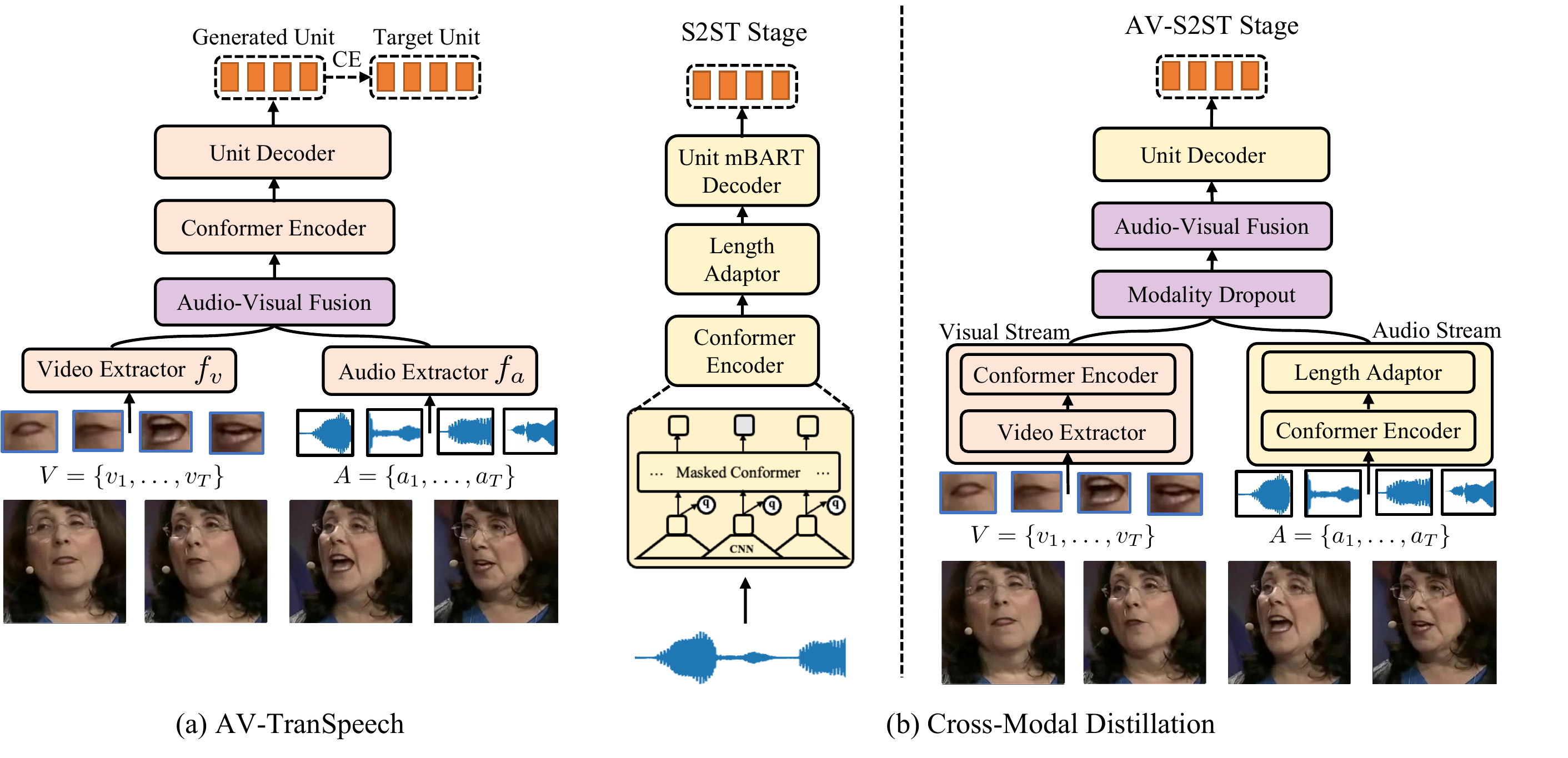}
   \caption{In subfigure (a), we compute the cross-entropy loss (denoted as "CE") during training. In subfigure (b), we initialize the visual and audio streams from AV-Hubert and S2ST models for cross-modal distillation in a low-resource setup. The modality dropout denoted with dotted lines is excluded during inference.} 
    \label{fig:arch}
  \end{figure*}

In this section, we first overview the encoder-decoder architecture for AV-TranSpeech, following which we introduce the cross-modal distillation procedure for few-shot transfer learning with low-resource data. The overall architecture has been illustrated in Figure~\ref{fig:pipeline}, and we put more details on the encoder and decoder block in Appendix~\ref{appendix:arch}.

\subsection{Overview}
The overall AV-S2ST pipeline has been illustrated in Figure~\ref{fig:pipeline}, where we 1) use the SSL HuBERT~\citep{hsu2021hubert} to derive discrete units of target speech; 2) build the audio-visual speech-to-unit translation (AV-S2UT) and 3) apply a separately trained unit-based vocoder to convert the translated units into waveform.

For audio-visual speech-to-unit translation, we adopt the encoder-decoder sequence-to-sequence model as the backbone. The audio-visual speech samples first pass through the multi-layer audio and video feature extractors, which are then fused and fed into the backbone conformer encoder. In the following, the unit decoder autoregressively predicts unit sequences corresponding to the target speech.

Training direct textless AV-S2ST models typically requires a large amount of parallel training data~\citep{duquenne2022speechmatrix,lee2021textless}, while resources providing parallel multimodal data could be limited due to the heavy workload. To alleviate the issue of data scarcity, we 1) build upon the recently introduced Audio-Visual HuBERT (AV-HuBERT) which learns contextual representations through self-supervised masked prediction, and show that large-scale pre-training benefits AV-S2ST training; 2) introduce the cross-modal distillation with S2ST models trained on the audio-only corpus, which further reduces the requirements of visual data and boosts the performance of visual systems in low-resource scenarios (10h, 30h).

\subsection{Pre-Trained Encoder}

Audio-Visual Hidden Unit BERT (AV-HuBERT) is a self-supervised model that learns from unlabeled audio-visual speech data. AV-HuBERT comprises four modules: a feed-forward network (FFN) audio feature extractor, a modified ResNet~\citep{stafylakis2017combining,martinez2020lipreading} video feature extractor, a fusion module, and a conformer~\citep{gulati2020conformer} backend. 

Denote the domain of visual and audio samples by $V, A \subset \mathbb{R}$ respectively. The source language is therefore a sequence of visual $V=\left\{v_{1}, \ldots, v_{T}\right\}$ and speech $A=\left\{a_{1}, \ldots, a_{T}\right\}$ samples for $T$ time-steps. The multi-layer audio feature extractor $f_a$ and the video feature extractor $f_v$ respectively take audio $A$ and visual frames $V$ as input, which are then fused (i.e., element-wise addition) and fed into the backbone conformer encoder and generates contextual representations $X=\left\{x_{1}, \ldots, x_{T}\right\}$.

\subsection{Unit Decoder}

The autoregressive decoder is assisted with an attention module, which takes the encoder output as the source values for the attention, and predicts unit sequences corresponding to the target translated speech. We use a stack of transformer layers as the decoder, along with a multi-head attention~\citep{vaswani2017attention}. Given the $T$-frame contextual representations from source speech $X=\left\{x_{1}, \ldots, x_{T}\right\}$, autoregressive model $\theta$ factors the distribution over possible outputs $Y=\left\{y_{1}, \ldots, y_{N}\right\}$ by:
\begin{equation}
    p(Y \mid X ; \theta)=\prod_{i=1}^{N+1} p(y_{i} \mid y_{0: i-1}, x_{1: T} ; \theta),
\end{equation} 
where the special tokens $y_{0} (\langle bos \rangle)$ and $y_{N+1} (\langle eos \rangle)$ are used to represent the beginning and end of all target units. 

\subsection{Cross-Modal Distillation}
In this part, we investigate the transfer learning from orders of magnitude audio data to boost the performance of visual systems. Specifically, we leverage the S2ST model trained on a large-scale audio-only corpus and perform \textbf{cross-modal distillation} with low-resource audio-visual data.

\textbf{S2ST Model.} We adopt the current state-of-the-art S2ST model~\citep{popuri2022enhanced} with pre-trained wav2vec 2.0~\citep{baevski2020wav2vec} and mBART~\citep{liu2020multilingual}. Wav2vec 2.0 is a self-supervised framework to learn speech representations from unlabeled audio data, which is trained via contrastive loss with masked spans on the input to the context encoder. mBART has been originally proposed for denoising autoencoder over text sequences, which predicts the original text $z$ given its noisy version, $g(z)$, created by random masking. As such, the powerful S2ST model provides a significant initialization for training audio-visual systems.

\textbf{Modality Adaptor.} The encoder in S2ST model $f_{a}$ encodes speech representation with $A=\left\{a_{1}, \ldots, a_{T'}\right\}$ with a stride of about 20ms, while the video stream $f_{v}$ generates visual feature sequence $V=\left\{v_{1}, \ldots, v_{T}\right\}$ at a stride of 10 ms from the raw waveform. To alleviate this length mismatch between the audio and visual representations, we add a randomly initialized modality adaptor layer consisting of a single 1-D convolutional layer with stride 2 between the audio and video streams.             

\textbf{Modality Dropout.} Since S2ST models provide a strong initialization for our AV-S2ST, and thus it can relate audio input to lexical output more effortlessly than the visual input stream, leading to the domination of audio modality in model decisions. To prevent the model’s over-reliance on the audio stream in our joint model, we include a modality dropout with $p=50\%$ probabilities to mask the full features of one modality before fusing audio and visual inputs, forcing the visual encoder to learn contextual representations. We show feature fusion in our cross-modal distillation with modality dropout:
\begin{equation}
    \nonumber
    \label{fusion}
    f_{av}= \begin{cases} f_{a} + f_{v} & \text { with } p=0.5 \\ f_{a} +  \mathbf{0} & \text { with } p=0.25 \\ \mathbf{0} + f_{v} & \text { with } p=0.25
    \end{cases}
\end{equation}
As such, modality dropout~\citep{chern2022audio,zhang2019robust} prevents the model from ignoring video input and encourages the model to produce the prediction regardless of what modalities are used as input.

\subsection{Model Training}
In training AV-TranSpeech, we compute the cross-entropy loss (denoted as "CE") between generated and reference units. For low-resource scenarios, we group our distillation strategies into two categories: 1) for AV-S2ST, we adapt the speech encoder and the unit decoder in S2ST to the visual system; 2) for V2ST with visual-only input, we only transfer the knowledge from unit decoder in S2ST to avoid noisy and incomplete decoding. In this way, the dependence on a large number of parallel audio-visual data can be reduced for constructing visual systems.

\section{Experiments}

\subsection{Experimental Setup}

Following the common practice in the direct unit S2ST pipeline, we apply the publicly-available pre-trained multilingual HuBERT (mHuBERT) model and unit-based HiFi-GAN vocoder~\citep{polyak2021speech,kong2020hifi}, leaving them unchanged. 

\subsubsection{Dataset} 
To evaluate the performance of the proposed model, we conduct experiments on two language pairs, including English-Spanish (En-Es), and English-French (En-Fr). 

\textbf{LRS3-T.} We construct our translation dataset by converting the transcribed English text from LRS3~\citep{afouras2018lrs3} into target language using cascaded neural machine translation (NMT) and text-to-speech (TTS) systems. We remove short clips (less than 2 seconds) and discard the non-vocal segments with voice activation detection (VAD). To this end, we collect 200-hour parallel audio-visual translation data (with source videos and target speech), namely LRS3-T which we plan to release.


\textbf{CVSS-C.} For training S2ST models, we use the benchmark dataset CVSS-C~\citep{jia2022cvss}, which is derived from the CoVoST 2~\citep{wang2020covost} speech-to-text translation corpus by synthesizing the translation text into speech using a single-speaker TTS system. 

\textbf{Noise.} For evaluating our AV-S2ST models under different noise categories, we prepare noise audio clips in the categories of ``music'' and ``babble'' sampled from MUSAN dataset~\citep{snyder2015musan}, and create ``speech'' noise samples following~\citet{popuri2022enhanced}.

The total duration of each dataset is shown in Table~\ref{table:dataset}.

\begin{table}[ht]
    \centering
    \small
    \begin{tabular}{c|c|c|c|c}
    \toprule
    Dataset                 & Subset & Modality            & En-Es      & En-Fr    \\
    \midrule
    \multirow{3}{*}{LRS3-T} & Normal                       & \multirow{3}{*}{AV} & \multicolumn{2}{c}{200} \\
                            & Small                        &                     & \multicolumn{2}{c}{30}  \\
                            & Tiny                        &                     & \multicolumn{2}{c}{10}  \\
    \midrule
    CVSS-C                    & /                          & A                   & 69.5       & 170        \\
    \midrule
    \multirow{3}{*}{Noise}  & Music                      & \multirow{3}{*}{A}  & \multicolumn{2}{c}{35}  \\
                            & Babble                     &                     & \multicolumn{2}{c}{20}  \\
                            & Speech                     &                     & \multicolumn{2}{c}{50}  \\
    \bottomrule
    \end{tabular}
    \caption{Total duration in hours of samples in different datasets.}
    \label{table:dataset}
    \end{table}

\begin{table*}[h]
    \small
    \centering
    \begin{tabular}{c|c|c|cc|cc}
    \toprule
    \multirow{2}{*}{ID} & \multirow{2}{*}{Pre-Training}   & \multirow{2}{*}{Modality} & \multicolumn{2}{c|}{En-Es} & \multicolumn{2}{c}{En-Fr} \\
                        &         &                   & BLEU & MOS  & BLEU & MOS  \\
    \midrule
     1  & /      & AV         &  0.67     & /                   &  1.01         &    /    \\
     2  & AVHubert & AV       &  45.2     &  3.82$\pm$0.09      &  33.6      &  3.98$\pm$0.08     \\
     \midrule                     
     3  & /         & A       &  0.51     &   /                 &  0.90     &   /  \\
     4  & AVHubert & A        &  43.1     &  3.80$\pm$0.09      &  31.6      &   3.90$\pm$0.07    \\
     \midrule                       
     5  & /        & V        &  0.18     &   /                 &  0.32      &  / \\
     6  & AVHubert & V        &  25.0     &   3.94$\pm$0.11     & 19.9      &    3.95$\pm$0.10    \\
     \midrule   
     7  & Enhanced S2ST & A   &  42.5     &   3.88$\pm$0.10     &  32.0    &   3.91$\pm$0.09     \\
      8  & Unit TTS  & /      &  67.7     &   4.04$\pm$0.07     &  54.1     &   4.09$\pm$0.10     \\
      9  & NMT+TTS   & /      &  76.0     &   4.15$\pm$0.08     &  63.9     &   4.20$\pm$0.10    \\
     \bottomrule
    \end{tabular}
    \caption{Translation quality (BLEU ($\uparrow$)) and speech naturalness (MOS ($\uparrow$)) comparison with baseline systems. We set the beam size to $10$ in autoregressive decoding.}
    \label{table:avs2st}
    \end{table*}

\subsubsection{Model Configurations and Training}
For training S2ST models, we adopt Wav2vec 2.0 LARGE pre-trained on Libri-light dataset~\citep{kahn2020libri} as audio encoder and unit mBART pre-trained on VoxPopuli dataset~\citep{wang2021voxpopuli} as the decoder. Following the practice in unit-based S2ST~\citep{lee2021direct}, we use the k-means algorithm to cluster the representation given by the normalized mHuBERT~\citep{huang2022transpeech} into a vocabulary of 1000 units as training targets. The inputs to AV-TranSpeech are lip Regions-Of-Interest (ROIs) for the visual stream and 80-dimensional mel-filterbank features at every 10-ms for the audio stream. As the image frames are sampled at 25Hz, we stack the 4 neighboring acoustic frames to synchronize the two modalities. The encoders in AV-TranSpeech follow AV-HuBERT LARGE configuration with 24 transformer blocks, each with 16 attention heads and 1024/4096 embedding/feed-forward dimensions. We remove the auxiliary tasks for simplification and follow the unwritten language scenario~\citep{lee2021textless}. AV-TranSpeech is trained until convergence for 20k steps using the Adam optimizer ($\beta_{1}=0.9, \beta_{2}=0.98, \epsilon=10^{-8}$) with 6 Tesla V100 GPU. A comprehensive table of hyperparameters is available in Appendix~\ref{appendix:arch}.

\begin{figure*}[ht]
    \vspace{-2mm}
    \centering
    \includegraphics[width=0.9\textwidth,trim={1.0cm 0cm 1.0cm 0cm}]{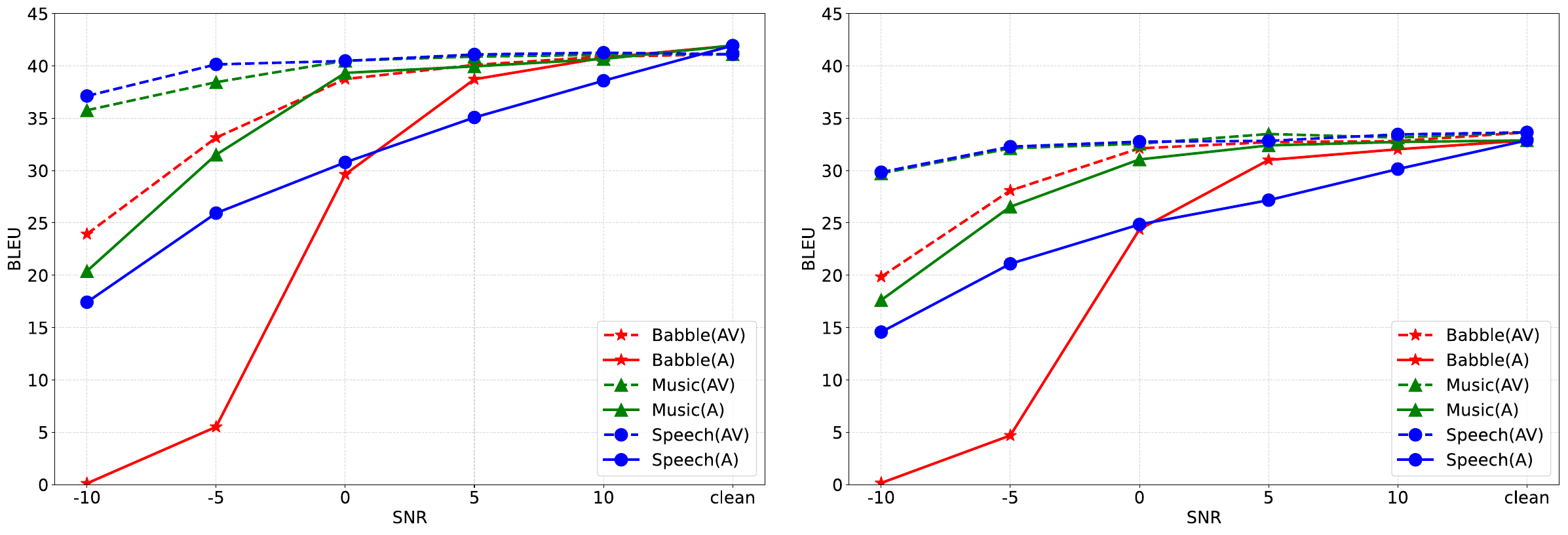}
   \caption{Illustration of model performance with different noise configurations and input modalities.} 
    \label{fig:noise}
  \end{figure*}

\subsubsection{Evaluation}
We compare AV-TranSpeech with other systems using the publicly-available \textit{fairseq} framework~\citep{ott2019fairseq}, including, 1) NMT+TTS cascaded system to simulate the construction of LRS3-T, where we adopt MMT to convert transcribed English text to target languages (regarded as \textit{reference text}) and then apply TTS model for speech generation, following we transcribe the speech and compute the BLEU; 2) Unit TTS, where we first synthesize speech samples with target units, and then transcribe the speech and compute BLEU; 3) Moreover, we compare the performance of AV-TranSpeech in S2ST with baseline model~\citep{popuri2022enhanced} (denoted as \textit{Enhanced S2ST}).

For translation accuracy, we use open-sourced ASR models in \textit{fairseq} framework to transcribe the audios and then calculate the BLEU score~\citep{papineni2002bleu} between the generated and the reference text. To evaluate the naturalness of the speech output, we conduct crowd-sourced human evaluations with MOS, rated from 1 to 5 and reported with 95\% confidence intervals (CI) via Amazon Mechanical Turk. 
More details on evaluation have been attached in Appendix~\ref{evaluation}.

\begin{table*}[ht]
    \small
    \centering
    \begin{tabular}{c|ccccc|ccccc|ccccc}
    \toprule
    \multirow{2}{*}{Modality}  & \multicolumn{5}{c|}{Babble (SNR)} & \multicolumn{5}{c|}{Music (SNR)} & \multicolumn{5}{c}{Speech (SNR)} \\
      & -10    & -5   & 0   & 5   & 10   & -10   & -5   & 0   & 5   & 10   & -10    & -5   & 0   & 5   & 10   \\
    \midrule        
    \multicolumn{16}{l}{\bfseries En-Es Translation}    \\
    \midrule 
AV      & 23.9 & 33.1 & 38.7 & 40.0 & 40.8 & 35.7 & 38.4 & 40.4 & 40.8 & 41.0 & 37.1 & 40.1 & 40.4 & 41.0 & 41.2     \\
A       & 0.1  & 5.5  & 29.6 & 38.7 & 40.7 & 20.3 & 31.5 & 39.3 & 39.9 & 40.6 & 17.4 & 25.9 & 30.7 & 35.0 & 38.5  \\
    \midrule        
    \multicolumn{16}{l}{\bfseries En-Fr Translation}    \\
    \midrule 
AV      & 19.8 & 28.0 & 32.1 & 32.7 & 32.8 & 29.7 & 32.1 & 32.5 & 33.4 & 33.5 & 29.8 & 32.2 & 32.7 & 32.8 & 33.4 \\
A       & 0.1  & 4.6  & 24.3 & 31.0 & 31.2 & 17.6 & 26.5 & 30.8 & 31.0 & 31.2 & 14.5 & 21.0 & 24.8 & 27.1 & 30.1 \\
    \bottomrule
    \end{tabular}
    \caption{Translation accuracy (BLEU scores ($\uparrow$)) comparison among models with different noise configurations and input modalities.}
    \label{table:noises2st}
    \end{table*}

\subsection{Translation Accuracy and Speech Naturalness}
Table~\ref{table:avs2st} summarizes the translation accuracy and speech naturalness among all systems, and we have the following observations: 1) \textbf{Large-scale multimodal pre-training (1 vs. 2)} improves performance by a large margin, while the naive model fails to work without the self-supervised pre-training strategy. It is mainly because LRS3-T is a challenging unconstrained dataset with a large proportion of videos collected from TED talks, showing the difficulty~\citep{zhang2020uwspeech,jia2019direct} of direct speech-to-speech translation without relying on intermediate texts or auxiliary multitask training. In contrast, with a pre-trained AVHubert encoder and a randomly initialized decoder, AV-TranSpeech is efficient in learning contextual representations from audio-visual signals. 2) \textbf{Visual modality (2 vs. 4)} has brought a gain of 2.0 BLEU points on average. It complements the audio stream with visual information, opening up a host of practical applications: enabling silent dictation or dubbing archival silent films. 3) We further compare AV-TranSpeech in S2ST with the baseline model (4 vs. 7), showing that AV-TranSpeech with audio-only input is on-par with the current state-of-the-art speech model in terms of translation accuracy. 4) For speech quality, AV-TranSpeech produces natural speech regardless of modalities input competitive with the baseline S2ST system. Since we apply the publicly-available pre-trained unit vocoder which mainly controls the naturalness of output speech and leave it unchanged, we expect AV-TranSpeech exhibits high-quality speech generation as baselines.

\begin{table}[ht]
    \centering
    \small
    \begin{tabular}{c|c|c|cc}
        \toprule
    ID & Audio & Modality & En-Es & En-Fr \\
    \midrule        
    \multicolumn{5}{l}{\bfseries Finetune with 10 hours data}    \\
    \midrule 
      1     & \multirow{2}{*}{/}      & AV          &  7.2    & 6.0      \\
      2     &                         & V           &  3.7    & 4.6      \\
   \midrule 
      3     & \multirow{2}{*}{Covost} & AV          &  21.5    & 24.4      \\
      4     &                         & V           &  5.8     & 6.4      \\
 \midrule        
 \multicolumn{5}{l}{\bfseries Finetune with 30 hours data}    \\
 \midrule 
     5      &  \multirow{2}{*}{/}      & AV        &  13.0  &  11.5   \\
     6      &                          & V         &  8.9   &  7.5    \\
 \midrule 
     7      &  \multirow{2}{*}{Covost} & AV        &  22.2  & 28.3    \\
    8       &                          & V         &  10.4 &  9.6     \\
    \bottomrule
    \end{tabular}
\caption{Leveraging audio-only data for boosting the performance of visual systems (AV or V) in low-resource scenarios. Audio: S2ST model trained on audio-only data.}
\label{table:fewshots2st}
\vspace{-3mm}
\end{table}

\begin{figure*}[ht]
    \vspace{-4mm}
    \centering
    \includegraphics[width=1.0\textwidth]{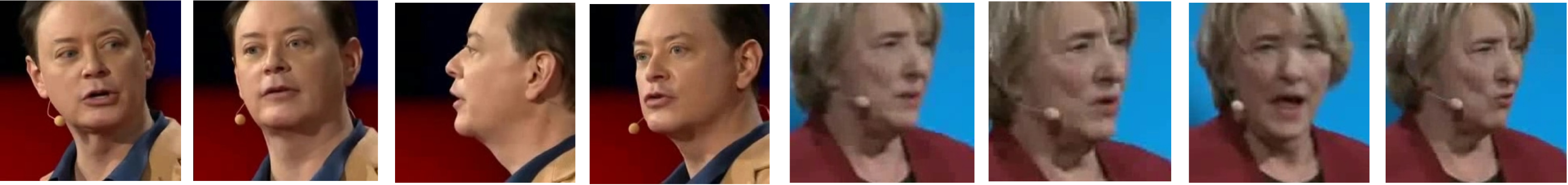}
  \end{figure*}

\begin{table*}[ht]
    \centering
    \small
    \begin{tabular}{l|l}
    \toprule
    Source:      & and the soldier on the front tank said we have unconditional orders to destroy this.                \\
    Target:      &  le soldat sur le char de front a dit que nous avions des ordres inconditionnels pour détruire cela.          \\
    V2ST:         & le soldat \textbf{a} dit que nous avons des ordres \textbf{conditionnels à} détruit pour détruire ce. \\
    S2ST:          & le soldat sur le \textbf{premier ban} a dit que nous avons des ordres \textbf{non} conditionnels pour détruire cela.            \\
    Noisy S2ST:    &  \textbf{nous avons également des ordres constamment} pour détruire cela. \\
    AV-S2ST:      & le soldat sur le premier \textbf{réservant} a dit que nous avons des ordres inconditionnels pour détruire cela.           \\
    Noisy AV-S2ST:  &   le soldat sur \textbf{ledexpert} a dit que nous avons des ordres \textbf{nom policières} pour détruire cela. \\
    \midrule
    Source:      & we met men where they were at and we built a program. \\
    Target:      & nous avons rencontré des hommes là où ils étaient et nous avons construit un programme. \\
    V2ST:         & nous avons rencontré des hommes \textbf{lorsquils sétaient agis et que} nous avons construit un programme. \\
    S2ST:          &  nous avons rencontré des hommes \textbf{quand} ils étaient \textbf{atteurs} et nous avons construit un programme.             \\
    Noisy S2ST:    & \textbf{intéressante de la classe de mathématiques et de leur données de maternelle.} \\
    AV-S2ST:      &  nous avons rencontré des hommes où ils étaient \textbf{atés} et nous avons construit un programme. \\
    Noisy AV-S2ST:    & nous avons rencontré des hommes \textbf{quand ils} étaient \textbf{atteurs} et nous avons construit un programme.  \\
    \bottomrule
    \end{tabular}
    \caption{Two examples comparing translations produced by AV-TranSpeech with different modalities. The left video frames refer to the first example. We use the bond fonts to indicate the the issue of \textbf{noisy and incomplete translation.} We use SNR=0 with babble noise for both noisy scenarios.}
    \label{table:cast}
    \vspace{-3mm}
    \end{table*}

\subsection{Visual Modality Evaluation}

The benefit of incorporating the visual stream is more apparent in challenging scenarios~\citep{afouras2018deep,popuri2022enhanced}, and thus we evaluate our models in the noisy setting to examine the impact of input modality (audio or audio-visual). A noise category with an audio clip has been sampled each time, following which we randomly mix the sampled noise with varied probabilities at five SNR levels: $\{-10, -5, 0, 5, 10\}$ dB. For easy comparison, the results are presented in Table~\ref{table:noises2st} and visualized in Figure~\ref{fig:noise}, and we have the following observations: 

1) AV-S2ST consistently outperforms audio-only S2ST under all settings regardless of the SNR and the type of noise. AV-TranSpeech complements the audio stream with visual information, which is invariant to speaking environments and promotes robustness. 2) As the volume of the noise increase with lower SNR, both languages have presented a degradation in translation accuracy. Informally, AV-S2ST models show a relatively slower BLEU drop (42\% drop in SNR-10 babble noise), while a distinct decrease could be witnessed in audio-only S2ST models (99.9\% drop in SNR-10 babble noise).


\subsection{Low Resource Evaluation}
Training direct AV-S2ST models without relying on intermediate text typically requires a large amount of parallel visual speech (i.e., lip) training data, while there may be very few resources due to the heavy workload. In this section, we prepare low-resource audio-visual data (LRS3-T 10h, 30h) and leverage large-scale audio-only data (Covost) to boost the performance of visual systems (AV-S2ST, V2ST), to investigate the effectiveness of our cross-modal distillation. The results are compiled and presented in Table~\ref{table:fewshots2st}, and we have the following observations: 

1) In consistent with previous practice~\citep{duquenne2022speechmatrix,tjandra2019speech}, training speech models are faced with the significant issue of data scarcity. As training data is reduced in the low-resource scenario, a distinct degradation in translation accuracy could be witnessed in both modalities (AV-S2ST or V2ST). 2) Leveraging orders of magnitude audio-only data with cross-modal distillation, the visual systems achieve BLEU scores of 21.5 and 22.2 respectively in En-Es and En-Fr AV-S2ST, showing a significant promotion regardless of the modalities and languages. In this way, the dependence on a large number of parallel audio-visual data can be reduced for constructing visual systems.



\subsection{Case Study}

We present several translation examples sampled from the En-Fr language pair in Table~\ref{table:cast}, and have the following findings: 1) With the complemental visual information brought in, the results produced by AV-TranSpeech are noticeably more literal. 2) Moreover in challenging noisy scenarios, S2ST models suffer severely from the issue of \textit{noisy and incomplete translation}, which is largely alleviated in AV-S2ST. AV-S2ST consistently outperforms audio-only S2ST in a noisy environment.



\section{Conclusion}

In this work, we proposed AV-TranSpeech, the first audio-visual speech-to-speech translation (AV-S2ST) model without relying on intermediate text. AV-TranSpeech complemented the audio stream with the visual information to promote robustness in noisy environments, opening up a host of practical applications: silent dictation or dubbing archival films. To mitigate the data scarcity for training AV-S2ST models, we 1) built upon the AV-HuBERT with a self-supervised learning framework fir contextual representations, showing that large-scale pre-training benefited the AV-S2ST training; 2) leveraged cross-modal distillation with S2ST models trained on the audio-only corpus, which further reduced the visual data requirements and boosted performance in low-resource scenarios. Experimental results on two language pairs demonstrated that AV-TranSpeech achieved significant robustness in noisy environments, outperforming audio-only S2ST models under all settings regardless of the type of noise. With low-resource audio-visual data (10h, 30h), cross-modal distillation yielded an improvement of 7.6 BLEU on average compared with baselines. We envisage that our work will serve as a basis for future audio-visual speech-to-speech translation studies, unlocking the ability for high-quality translation given a user-defined modality input.

\clearpage

\section{Limitation and Potential Risks}
As mentioned in our experimental setup, we provide results of AV-S2ST in LRS3-T with synthesized target speech, similar to the pioneer literature~\citep{jia2022cvss} in S2ST. One of our future directions is to develop a better benchmark dataset (e.g., mined or human-annotated data) to improve translation performance.

As mentioned in our results analysis, the BLEU scores heavily depend on the ASR quality, which may not accurately reflect the speech translation performance. Future directions could be improving ASR quality or exploring other evaluation metrics without reliance on ASR models.

AV-TranSpeech lowers the requirements for audio-visual speech-to-speech translation, which may cause unemployment for people with related occupations such as interpreter and translator. In addition, there is the potential for harm from non-consensual voice generation or fake media. The voices of the speakers in the recordings might be overused than they expect.

\section*{Acknowledgements}

This work was supported in part by the National Key R\&D Program of China under Grant No.2022ZD0162000, National Natural Science Foundation of China under Grant No.62222211, Grant No.61836002 and Grant No.62072397.

\bibliography{custom}
\bibliographystyle{acl_natbib}

\clearpage
\appendix

\begin{table*}[h]
  \small
  \centering
  \begin{tabular}{l|c|c}
  \toprule
  \multicolumn{2}{c|}{Hyperparameter}   & AV-TranSpeech \\ 
  \midrule
  \multirow{4}{*}{Conformer Encoder} 
  &Encoder Layer                     &    24     \\
  &Encoder Input/Output Dim           &  1024      \\  
  &Encoder FFN Embed Dim              &   4096 \\                       
  &Encoder Attention Heads           &  16      \\    
  &Encoder Dropout                   &  0.1   \\ 
  \midrule
  \multirow{3}{*}{Length Adaptor}       
  & Conv1d Layer       &  1      \\ 
  & Conv1d Kernel       &  3      \\ 
  & Conv1d Stride       &  3      \\ 
  \midrule
  \multirow{6}{*}{Unit Decoder}      
  &Decoder Layer                      &   12 \\    
  &Decoder Input/Output Dim            &   1024  \\    
  &Decoder FFN Embed Dim              &   4096 \\   
  &Decoder Attention Headers           &  16   \\       
  &Decoder Dropout                     &  0.1  \\     
  \midrule 
  \multicolumn{2}{c|}{Total Number of Parameters}   & 827 M  \\
  \bottomrule
  \end{tabular}
  \vspace{2mm}
  \caption{Hyperparameters of AV-TranSpeech.}
  \label{tab:hyperparameters}
  \end{table*}

  \begin{figure*}[h]
    \centering
    \includegraphics[width=0.9\textwidth]{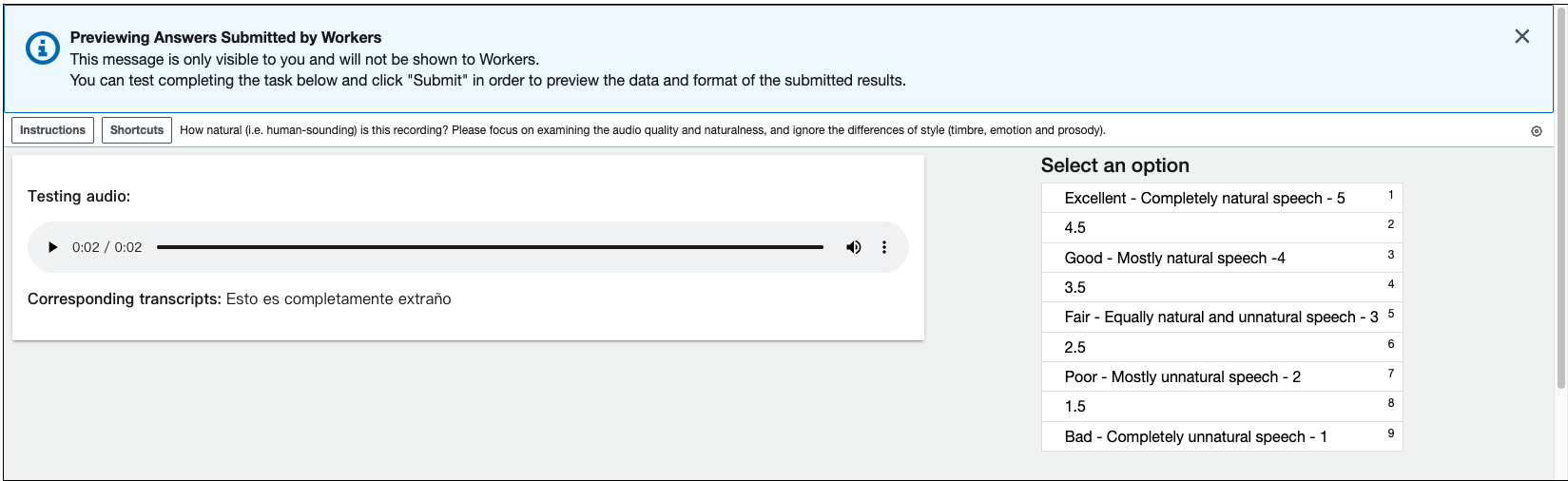}
   \caption{Screenshot of MOS testing.} 
    \label{fig:MOS}
   \end{figure*}

\section{Model Architectures} \label{appendix:arch}

In this section, we list the model hyper-parameters of AV-TranSpeech in Table~\ref{tab:hyperparameters}.

\section{Subjective Evaluation} 
\label{evaluation}

Following~\citep{huang2021multi,huang2022singgan}, all our Mean Opinion Score (MOS) tests are crowd-sourced and conducted by native speakers. The scoring criteria have been included in Table~\ref{matrix:naturalness} for completeness. The samples are presented and rated one at a time by the testers, each tester is asked to evaluate the subjective naturalness of a sentence on a 1-5 Likert scale. The screenshots of instructions for testers are shown in Figure~\ref{fig:MOS}. We paid \$8 to participants hourly and totally spent about \$500 on participant compensation.

\begin{table}[H]
 \centering
\small
    \caption{Ratings that have been used in the evaluation of speech naturalness of synthetic and ground truth samples.}
    \vspace{2mm}
  \begin{tabular}{ccc}
  \toprule
  Rating & Naturalness & Definition                           \\
  \midrule
  1      & Bad        &  Very annoying and objectionable dist. \\
  2      & Poor       &  Annoying but not objectionable dist. \\
  3      & Fair       &  Perceptible and slightly annoying dist\\
  4      & Good       & Just perceptible but not annoying dist. \\
  5      & Excellent  & Imperceptible distortions\\
  \bottomrule
  \end{tabular}
  \label{matrix:naturalness}
  \end{table}

\end{document}